\documentclass[10pt,twocolumn,letterpaper]{article}

\usepackage{cvpr}
\usepackage{times}
\usepackage{epsfig}
\usepackage{graphicx}
\usepackage{amsmath}
\usepackage{amssymb}
\usepackage{appendix}

\usepackage[pagebackref=true,breaklinks=true,letterpaper=true,colorlinks,bookmarks=false]{hyperref}

\cvprfinalcopy 

\def\cvprPaperID{2784} 

\ifcvprfinal\fi
\begin{document}

\title{Path Aggregation Network for Instance Segmentation} 
\author{Shu Liu$^{\dag}$
\and
Lu Qi$^{\dag}$
\and
Haifang Qin$^{\S}$
\and
Jianping Shi$^{\ddagger}$
\and
Jiaya Jia$^{\dag, \flat}$
\\
\and
$^{\dag}$The Chinese University of Hong Kong~~~~~~$^{\S}$Peking University\\
$^{\ddagger}$SenseTime Research~~~~~~$^{\flat}$YouTu Lab, Tencent\\
\vspace{-2mm}
{\tt\small \{sliu, luqi, leojia\}@cse.cuhk.edu.hk~~qhfpku@pku.edu.cn~~shijianping@sensetime.com} 
}
\maketitle

\begin{abstract}
The way that information propagates in neural networks is of great importance. In this paper, we propose Path Aggregation Network (PANet) aiming at boosting information flow in proposal-based instance segmentation framework. Specifically, we enhance the entire feature hierarchy with accurate localization signals in lower layers by bottom-up path augmentation, which shortens the information path between lower layers and topmost feature. We present adaptive feature pooling, which links feature grid and all feature levels to make useful information in each feature level propagate directly to following proposal subnetworks. A complementary branch capturing different views for each proposal is created to further improve mask prediction.

These improvements are simple to implement, with subtle extra computational overhead. Our PANet reaches the $1^{st}$ place in the COCO 2017 Challenge Instance Segmentation task and the $2^{nd}$ place in Object Detection task without large-batch training. It is also state-of-the-art on MVD and Cityscapes. Code is available at \url{https://github.com/ShuLiu1993/PANet}.

\end{abstract}

\section{Introduction} \label{sec:intro}
Instance segmentation is one of the most important and challenging tasks. It aims to predict class label and pixel-wise instance mask to localize varying numbers of instances presented in images. This task widely benefits autonomous vehicles, robotics, video surveillance, to name a few.

With the help of deep convolutional neural networks, several frameworks for instance segmentation, \eg, \cite{He2017Mask,li2017fully,bai2016deep,liu2017sgn}, were proposed where performance grows rapidly \cite{everingham2010pascal}. Mask R-CNN \cite{He2017Mask} is a simple and effective system for instance segmentation. Based on Fast/Faster R-CNN \cite{girshick2015fast,ren2015faster}, a fully convolutional network (FCN) is used for mask prediction, along with box regression and classification. To achieve high performance, feature pyramid network (FPN) \cite{lin2016feature} is utilized to extract in-network feature hierarchy, where a top-down path with lateral connections is augmented to propagate semantically strong features.

Several newly released datasets \cite{lin2014microsoft,cordts2016the,Neuhold2017the} make large room for algorithm improvement. COCO \cite{lin2014microsoft} consists of 200k images. Lots of instances with complex spatial layout are captured in each image. Differently, Cityscapes \cite{cordts2016the} and MVD \cite{Neuhold2017the} provide street scenes with a large number of traffic participants in each image. Blur, heavy occlusion and extremely small instances appear in these datasets. 

There have been several principles proposed for designing networks in image classification that are also effective for object recognition. For example, shortening information path and easing information propagation by clean residual connection \cite{he2016deep,he2016identity} and dense connection \cite{huang2016densely} are useful. Increasing the flexibility and diversity of information paths by creating parallel paths following the {\it split-transform-merge} strategy \cite{xie2016aggregated,chen2017dual} is also beneficial. 

\vspace{-0.1in}
\paragraph{Our Findings} Our research indicates that information propagation in state-of-the-art Mask R-CNN can be further improved. Specifically, features in low levels are helpful for large instance identification. But there is a long path from low-level structure to topmost features, increasing difficulty to access accurate localization information. Further, each proposal is predicted based on feature grids pooled from one feature level, which is assigned heuristically. This process can be updated since information discarded in other levels may be helpful for final prediction. Finally, mask prediction is made on a single view, losing the chance to gather more diverse information.

\begin{figure*}[bpt]
\centering
\begin{tabular}{@{\hspace{0mm}}c}
\includegraphics[width=0.9\linewidth]{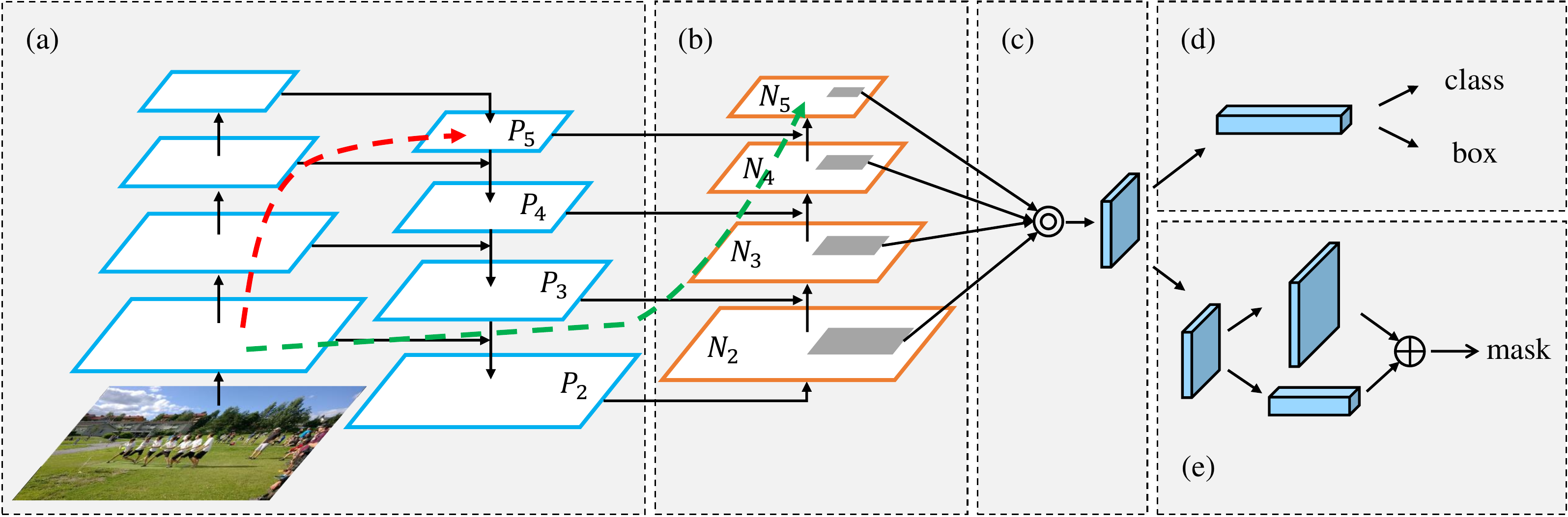}
\\
\end{tabular}
\caption{Illustration of our framework. (a) FPN backbone. (b) Bottom-up path augmentation. (c) Adaptive feature pooling. (d) Box branch. (e) Fully-connected fusion. Note that we omit channel dimension of feature maps in (a) and (b) for brevity.}
\label{fig:framework}
\vspace{-3mm}
\end{figure*}

\vspace{-0.1in}
\paragraph{Our Contributions} Inspired by these principles and observations, we propose PANet, illustrated in Figure \ref{fig:framework}, for instance segmentation. 

First, to shorten information path and enhance feature pyramid with accurate localization signals existing in low-levels, bottom-up path augmentation is created. In fact, features in low-layers were utilized in the systems of \cite{long2015fully,liu2016ssd,fu2017deconvolutional,noh2015learning,lin2016feature,cai2016a,kong2016hypernet,ghiasi2016laplacian}. But propagating low-level features to enhance entire feature hierarchy for instance recognition was not explored.

Second, to recover broken information path between each proposal and all feature levels, we develop adaptive feature pooling. It is a simple component to aggregate features from all feature levels for each proposal, avoiding arbitrarily assigned results. With this operation, cleaner paths are created compared with those of \cite{bell2016inside,zagoruyko2016a}. 

Finally, to capture different views of each proposal, we augment mask prediction with tiny fully-connected ({\it fc}) layers, which possess complementary properties to FCN originally used by Mask R-CNN. By fusing predictions from these two views, information diversity increases and masks with better quality are produced. 

The first two components are shared by both object detection and instance segmentation, leading to much enhanced performance of both tasks.

\vspace{-0.1in}
\paragraph{Experimental Results} With PANet, we achieve state-of-the-art performance on several datasets. With ResNet-50 \cite{he2016deep} as the initial network, our PANet tested with a single scale already outperforms champion of COCO 2016 Challenge in both object detection \cite{huang2017speed} and instance segmentation \cite{li2017fully} tasks. Note that these previous results are achieved by larger models \cite{he2016deep,szegedy2017inception-v4} together with multi-scale and horizontal flip testing. 

We achieve the $1^{st}$ place in COCO 2017 Challenge Instance Segmentation task and the $2^{nd}$ place in Object Detection task without large-batch training. We also benchmark our system on Cityscapes and MVD, which similarly yields top-ranking results, manifesting that our PANet is a very practical and top-performing framework. Our code and models are available at \url{https://github.com/ShuLiu1993/PANet}.

\section{Related Work}
\paragraph{Instance Segmentation}
There are mainly two streams of methods in instance segmentation. The most popular one is proposal-based. Methods in this stream have a strong connection to object detection. In R-CNN \cite{girshick2014rich}, object proposals from \cite{uijlings2013selective,zitnick2014edge} were fed into the network to extract features for classification. While Fast/Faster R-CNN \cite{girshick2015fast,ren2015faster} and SPPNet \cite{he2015spatial} sped up the process by pooling features from global feature maps. Earlier work \cite{hariharan2014simultaneous,hariharan2015hypercolumns} took mask proposals from MCG \cite{arbelaez2014multiscale} as input to extract features while CFM \cite{dai2015convolutional}, MNC \cite{dai2015instance} and Hayder \etal \cite{hayder2017boundary-aware} merged feature pooling to network for faster speed. Newer design was to generate instance masks in networks as proposal \cite{deepmask,sharpmask,dai2016instance-sensitive} or final result \cite{dai2015instance,liang2016reversible,liu2016multi}. Mask R-CNN \cite{He2017Mask} is an effective framework falling in this stream. Our work is built on Mask R-CNN and improves it from different aspects. 

Methods in the other stream are mainly segmentation-based. They learned specially designed transformation \cite{bai2016deep,li2017fully,liu2017sgn,uhrig2016pixel-level} or instance boundaries \cite{kirillov2017instancecut}. Then instance masks were decoded from predicted transformation. Instance segmentation by other pipelines also exists. DIN \cite{arnab2017pixelwise} fused predictions from object detection and semantic segmentation systems. A graphical model was used in \cite{zhang2015monocular,zhang2016instance} to infer the order of instances. RNN was utilized in \cite{paredes2016recurrent,ren2017end-to-end} to propose one instance in each time step.

\vspace{-0.1in}
\paragraph{Multi-level Features}
Features from different layers were used in image recognition. SharpMask \cite{sharpmask}, Peng \etal \cite{peng2017large} and LRR \cite{ghiasi2016laplacian} fused feature maps for segmentation with finer details. FCN \cite{long2015fully}, U-Net \cite{Ronneberger2015unet} and Noh \etal \cite{noh2015learning} fused information from lower layers through skip-connections. Both TDM \cite{Shrivastava2016beyond} and FPN \cite{lin2016feature} augmented a top-down path with lateral connections for object detection. Different from TDM, which took the fused feature map with the highest resolution to pool features, SSD \cite{liu2016ssd}, DSSD \cite{fu2017deconvolutional}, MS-CNN \cite{cai2016a} and FPN \cite{lin2016feature} assigned proposals to appropriate feature levels for inference. We take FPN as a baseline and much enhance it. 

ION \cite{bell2016inside}, Zagoruyko \etal \cite{zagoruyko2016a}, Hypernet \cite{kong2016hypernet} and Hypercolumn \cite{hariharan2015hypercolumns} concatenated feature grids from different layers for better prediction. But a sequence of operations, \ie, normalization, concatenation and dimension reduction are needed to get feasible new features. In comparison, our design is much simpler. 

Fusing feature grids from different sources for each proposal was also utilized in \cite{ren2017object}. But this method extracted feature maps on input with different scales and then conducted feature fusion (with the max operation) to improve feature selection from the input image pyramid. In contrast, our method aims at utilizing information from all feature levels in the in-network feature hierarchy with single-scale input. End-to-end training is enabled. 

\vspace{-0.1in}
\paragraph{Larger Context Region}
Methods of \cite{gidaris2015object,zeng2016crafting,zagoruyko2016a} pooled features for each proposal with a foveal structure to exploit context information from regions with different resolutions. Features pooled from a larger region provide surrounding context. Global pooling was used in PSPNet \cite{zhao2017pyramid} and ParseNet \cite{liu2015parsenet} to greatly improve quality of semantic segmentation. Similar trend was observed by Peng \etal \cite{peng2017large} where global convolutionals were utilized. Our mask prediction branch also supports accessing global information. But the technique is completely different.

\section{Framework}
Our framework is illustrated in Figure \ref{fig:framework}. Path augmentation and aggregation is conducted for improving performance. A bottom-up path is augmented to make low-layer information easier to propagate. We design adaptive feature pooling to allow each proposal to access information from all levels for prediction. A complementary path is added to the mask-prediction branch. This new structure leads to decent performance. Similar to FPN, the improvement is independent of the CNN structure, \eg, \cite{simonyan2014very,krizhevsky2012imagenet,he2016deep}.

\subsection{Bottom-up Path Augmentation}
\paragraph{Motivation}
The insightful point \cite{zeiler2014visualizing} that neurons in high layers strongly respond to entire objects while other neurons are more likely to be activated by local texture and patterns manifests the necessity of augmenting a top-down path to propagate semantically strong features and enhance all features with reasonable classification capability in FPN.

\begin{figure}[bpt]
\centering
\includegraphics[width=0.6\linewidth]{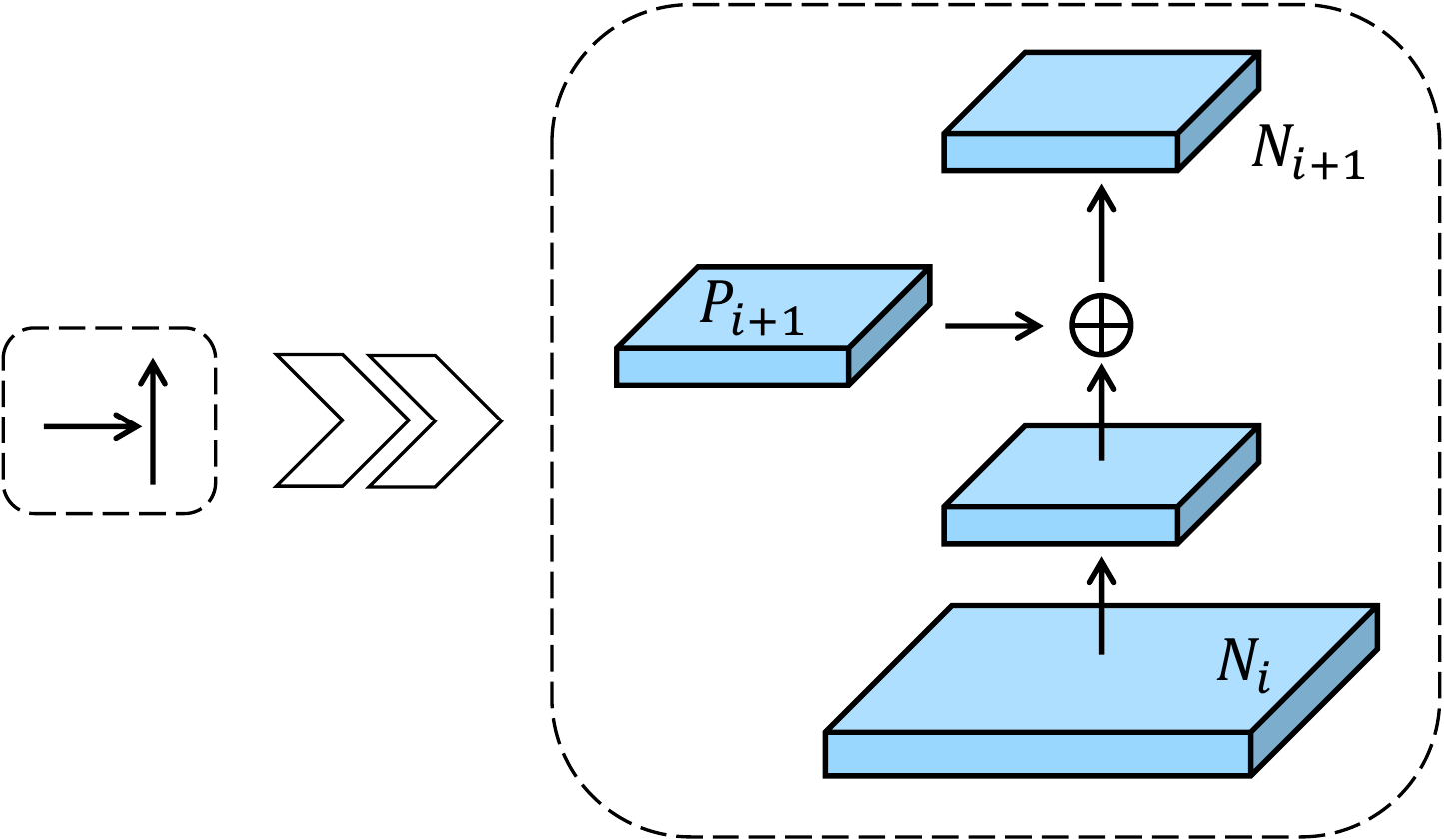}
\caption{Illustration of our building block of bottom-up path augmentation.}
\label{fig:bottom_up_block}
\vspace{-3mm}
\end{figure}

Our framework further enhances the localization capability of the entire feature hierarchy by propagating strong responses of low-level patterns based on the fact that high response to edges or instance parts is a strong indicator to accurately localize instances. To this end, we build a path with clean lateral connections from the low level to top ones. Therefore, there is a ``shortcut'' (dashed green line in Figure \ref{fig:framework}), which consists of less than $10$ layers, across these levels. In comparison, the CNN trunk in FPN gives a long path (dashed red line in Figure \ref{fig:framework}) passing through even $100$+ layers from low layers to the topmost one. 

\vspace{-0.1in}
\paragraph{Augmented Bottom-up Structure}
Our framework first accomplishes {\it bottom-up path augmentation}. We follow FPN to define that layers producing feature maps with the same spatial size are in the same network {\it stage}. Each feature level corresponds to one stage. We also take ResNet \cite{he2016deep} as the basic structure and use $\{P_2, P_3, P_4, P_5\}$ to denote feature levels generated by FPN. Our augmented path starts from the lowest level $P_2$ and gradually approaches $P_5$ as shown in Figure \ref{fig:framework}(b). From $P_2$ to $P_5$, the spatial size is gradually down-sampled with factor $2$. We use $\{N_2, N_3, N_4, N_5\}$ to denote newly generated feature maps corresponding to $\{P_2, P_3, P_4, P_5\}$. Note that $N_2$ is simply $P_2$, without any processing.

As shown in Figure \ref{fig:bottom_up_block}, each building block takes a higher resolution feature map $N_i$ and a coarser map $P_{i+1}$ through lateral connection and generates the new feature map $N_{i+1}$. Each feature map $N_i$ first goes through a $3 \times 3$ convolutional layer with stride $2$ to reduce the spatial size. Then each element of feature map $P_{i+1}$ and the down-sampled map are added through lateral connection. The fused feature map is then processed by another $3 \times 3$ convolutional layer to generate $N_{i+1}$ for following sub-networks. This is an iterative process and terminates after approaching $P_5$. In these building blocks, we consistently use channel $256$ of feature maps. All convolutional layers are followed by a ReLU \cite{krizhevsky2012imagenet}. The feature grid for each proposal is then pooled from new feature maps, \ie, $\{N_2, N_3, N_4, N_5\}$.

\subsection{Adaptive Feature Pooling}
\paragraph{Motivation}
In FPN \cite{lin2016feature}, proposals are assigned to different feature levels according to the size of proposals. It makes small proposals assigned to low feature levels and large proposals to higher ones. Albeit simple and effective, it still could generate non-optimal results. For example, two proposals with 10-pixel difference can be assigned to different levels. In fact, these two proposals are rather similar. 

Further, importance of features may not be strongly correlated to the levels they belong to. High-level features are generated with large receptive fields and capture richer context information. Allowing small proposals to access these features better exploits useful context information for prediction. Similarly, low-level features are with many fine details and high localization accuracy. Making large proposals access them is obviously beneficial. With these thoughts, we propose pooling features from all levels for each proposal and fusing them for following prediction. We call this process {\it adaptive feature pooling}. 

\begin{figure}[bpt]
\centering
\begin{tabular}{@{\hspace{0mm}}c}
\includegraphics[width=0.8\linewidth]{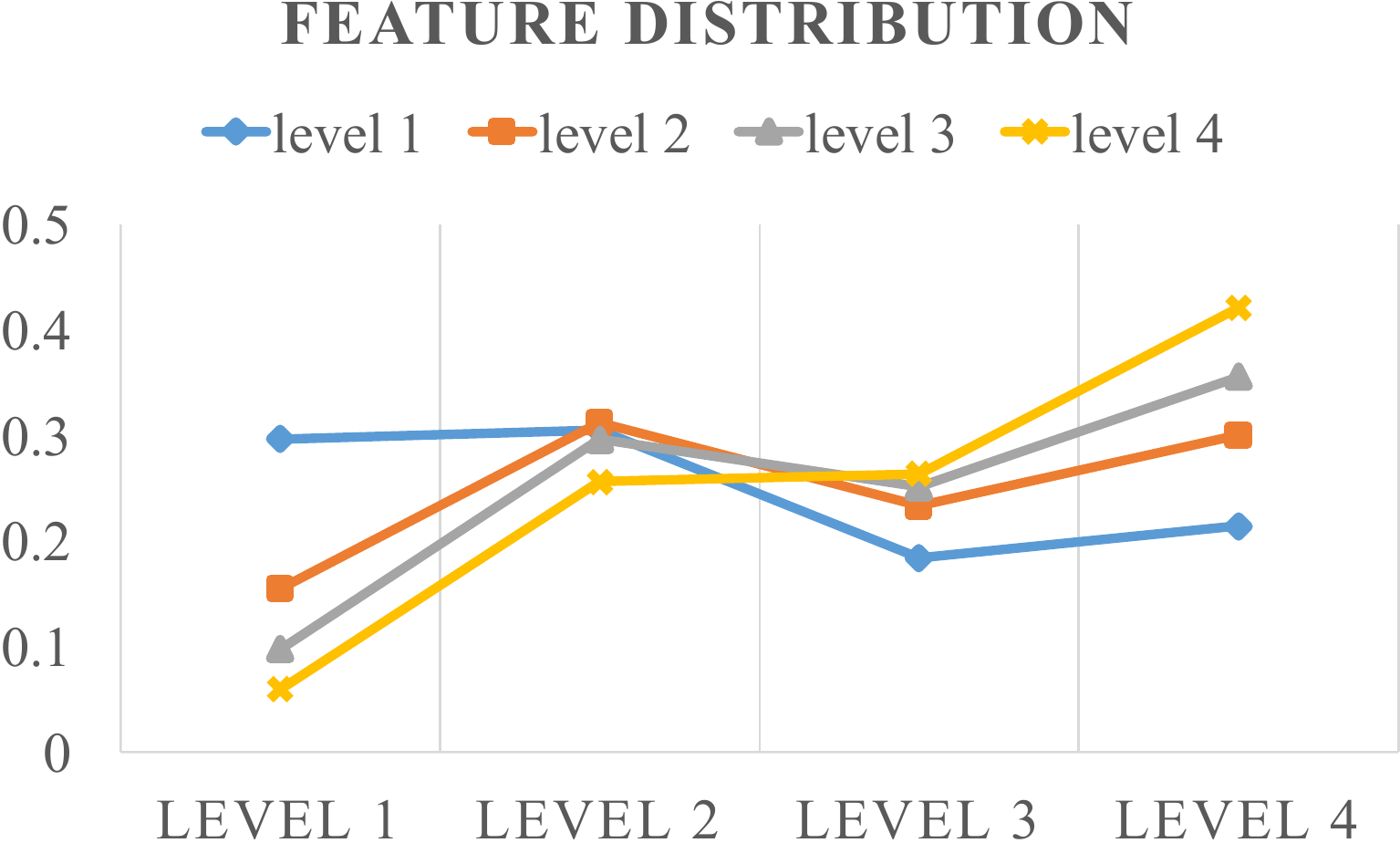}
\\
\end{tabular}
\caption{Ratio of features pooled from different feature levels with adaptive feature pooling. Each line represents a set of proposals that should be assigned to the same feature level in FPN, \ie, proposals with similar scales. The horizontal axis denotes the source of pooled features. It shows that proposals with different sizes all exploit features from several different levels.}
\label{fig:statistics}
\vspace{-0.1in}
\end{figure}

We now analyze the ratio of features pooled from different levels with adaptive feature pooling. We use {\it max} operation to fuse features from different levels, which lets network select element-wise useful information. We cluster proposals into four classes based on the levels they were assigned to originally in FPN. For each set of proposals, we calculate the ratio of features selected from different levels. In notation, levels $1 - 4$ represent low-to-high levels. As shown in Figure \ref{fig:statistics}, the blue line represents small proposals that were assigned to level $1$ originally in FPN. Surprisingly, nearly $70\%$ of features are from other higher levels. We also use the yellow line to represent large proposals that were assigned to level $4$ in FPN. Again, $50\%+$ of the features are pooled from other lower levels. This observation clearly indicates that {\it features in multiple levels together are helpful for accurate prediction}. It is also a strong support of designing bottom-up path augmentation.

\vspace{-0.1in}
\paragraph{Adaptive Feature Pooling Structure}
{\it Adaptive feature pooling} is actually simple in implementation and is demonstrated in Figure \ref{fig:framework}(c). First, for each proposal, we map them to different feature levels, as denoted by dark grey regions in Figure \ref{fig:framework}(b). Following Mask R-CNN \cite{He2017Mask}, ROIAlign is used to pool feature grids from each level. Then a fusion operation (element-wise max or sum) is utilized to fuse feature grids from different levels. 

In following sub-networks, pooled feature grids go through one parameter layer independently, which is followed by the fusion operation, to enable network to adapt features. For example, there are two {\it fc} layers in the box branch in FPN. We apply the fusion operation after the first layer. Since four consecutive convolutional layers are used in mask prediction branch in Mask R-CNN, we place fusion operation between the first and second convolutional layers. Ablation study is given in Section \ref{sec:afpabla}. The fused feature grid is used as the feature grid of each proposal for further prediction, \ie, classification, box regression and mask prediction. A detailed illustration of adaptive feature pooling on box branch is shown by Figure \ref{fig:ada} in Appendix.

Our design focuses on fusing information from in-network feature hierarchy instead of those from different feature maps of input image pyramid \cite{ren2017object}. It is simpler compared with the process of \cite{bell2016inside,zagoruyko2016a,kong2016hypernet}, where L-2 normalization, concatenation and dimension reduction are needed.

\subsection{Fully-connected Fusion}
\paragraph{Motivation}
Fully-connected layers, or MLP, were widely used in mask prediction in instance segmentation \cite{dai2015instance,liu2016multi,liang2016reversible} and mask proposal generation \cite{deepmask,sharpmask}. Results of \cite{dai2016instance-sensitive,li2017fully} show that FCN is also competent in predicting pixel-wise masks for instances. Recently, Mask R-CNN \cite{He2017Mask} applied a tiny FCN on the pooled feature grid to predict corresponding masks avoiding competition between classes. 

We note {\it fc} layers yield different properties compared with FCN where the latter gives prediction at each pixel based on a local receptive field and parameters are shared at different spatial locations. Contrarily, {\it fc} layers are location sensitive since predictions at different spatial locations are achieved by varying sets of parameters. So they have the ability to adapt to different spatial locations. Also prediction at each spatial location is made with global information of the entire proposal. It is helpful to differentiate instances \cite{deepmask} and recognize separate parts belonging to the same object. Given properties of {\it fc} and convolutional layers different from each other, we fuse predictions from these two types of layers for better mask prediction.

\begin{figure}[bpt]
\centering
\begin{tabular}{@{\hspace{0mm}}c}
\includegraphics[width=0.9\linewidth]{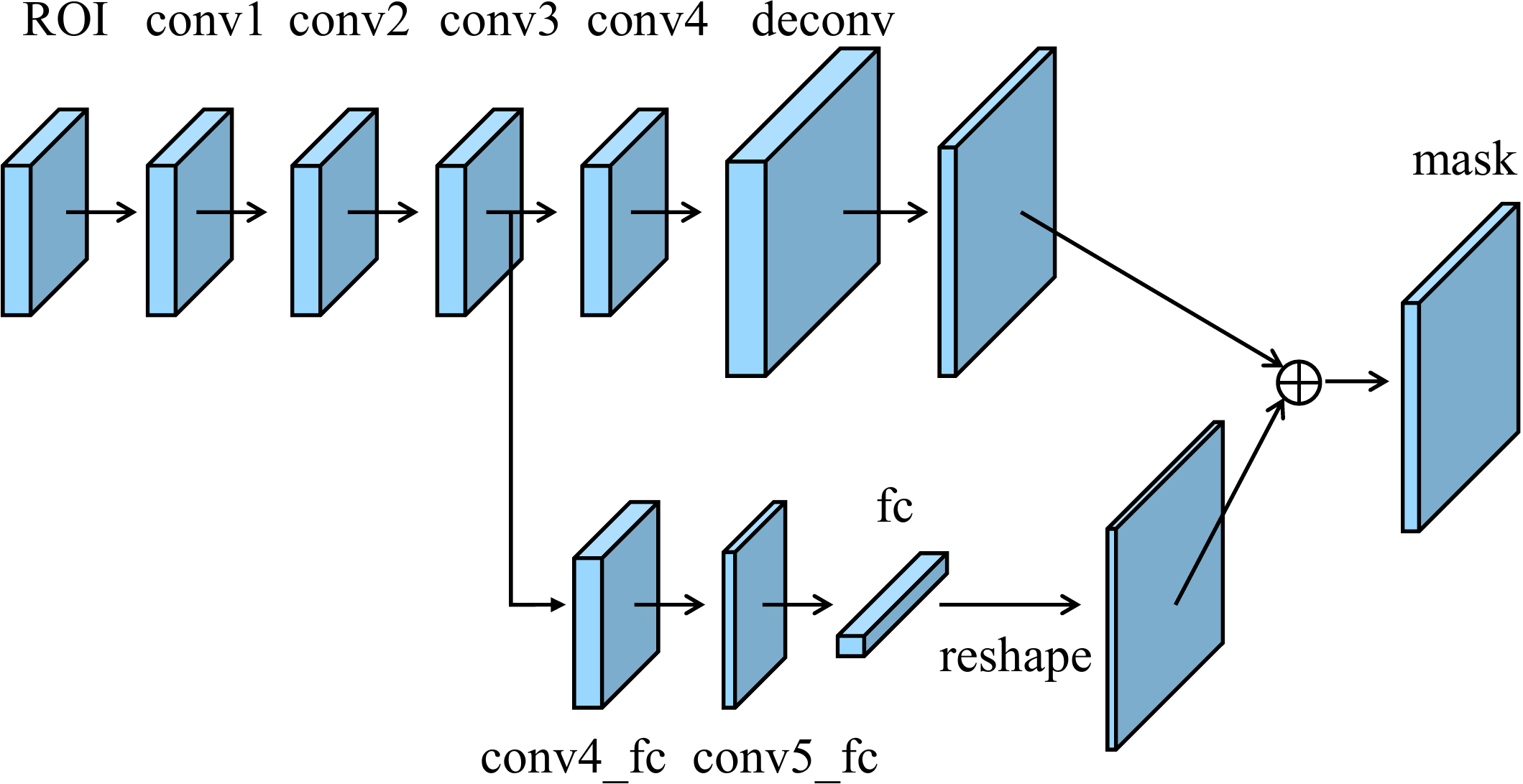}
\\
\end{tabular}
\caption{Mask prediction branch with fully-connected fusion.}
\label{fig:fc_aug}
\vspace{-0.1in}
\end{figure}

\begin{table*}[t]
	\centering 
	\footnotesize
	\begin{tabular}{c|ccc|ccc|c}
		Method & AP & AP$_{50}$ & AP$_{75}$ & AP$_S$ & AP$_M$ & $AP_L$ & Backbone \\
		\hline
		\hline
		Champion 2016 \cite{li2017fully} & 37.6 & 59.9 & 40.4 & 17.1 & 41.0 & 56.0 & $6\times$ResNet-101\\
		Mask R-CNN \cite{He2017Mask}+FPN \cite{lin2016feature} & 35.7 & 58.0 & 37.8 & 15.5 & 38.1 & 52.4 & ResNet-101 \\
		Mask R-CNN \cite{He2017Mask}+FPN \cite{lin2016feature} & 37.1 & 60.0 & 39.4 & 16.9 & 39.9 & 53.5 & ResNeXt-101 \\
		\hline
		\hline
		PANet / PANet [ms-train] & 36.6 / 38.2 & 58.0 / 60.2 & 39.3 / 41.4 & 16.3 / 19.1 & 38.1 / 41.1 & 53.1 / 52.6 & ResNet-50 \\
		PANet / PANet [ms-train] & 40.0 / \bf 42.0 & 62.8 / \bf 65.1 & 43.1 / \bf 45.7 & 18.8 / \bf 22.4 & 42.3 / \bf 44.7 & 57.2 / \bf 58.1 & ResNeXt-101 \\
	\end{tabular}\vspace{0.1cm}
	\caption{Comparison among PANet, winner of COCO 2016 instance segmentation challenge, and Mask R-CNN on COCO {\it test-dev} subset in terms of Mask AP, where the latter two are baselines. }\label{tab:maincoco}
	\vspace{-0.1in}

\end{table*}

\vspace{-0.1in}
\paragraph{Mask Prediction Structure}
Our component of mask prediction is light-weighted and easy to implement. The mask branch operates on pooled feature grid for each proposal. As shown in Figure \ref{fig:fc_aug}, the main path is a small FCN, which consists of $4$ consecutive convolutional layers and $1$ deconvolutional layer. Each convolutional layer consists of $256$ $3\times3$ filters and the deconvolutional layer up-samples feature with factor $2$. It predicts a binary pixel-wise mask for each class independently to decouple segmentation and classification, similar to that of Mask R-CNN. We further create a short path from layer $conv3$ to a {\it fc} layer. There are two $3\times3$ convolutional layers where the second shrinks channels to half to reduce computational overhead. 

A {\it fc} layer is used to predict a class-agnostic foreground/background mask. It not only is efficient, but also allows parameters in the {\it fc} layer trained with more samples, leading to better generality. The mask size we use is $28\times28$ so that the {\it fc} layer produces a $784\times1\times1$ vector. This vector is reshaped to the same spatial size as the mask predicted by FCN. To obtain the final mask prediction, mask of each class from FCN and foreground/background prediction from {\it fc} are added. Using only one {\it fc} layer, instead of multiple of them, for final prediction prevents the issue of collapsing the hidden spatial feature map into a short feature vector, which loses spatial information.

\section{Experiments}
We compare our method with state-of-the-arts on challenging COCO \cite{lin2014microsoft}, Cityscapes \cite{cordts2016the} and MVD \cite{Neuhold2017the} datasets. Our results are top ranked in all of them. Comprehensive ablation study is conducted on the COCO dataset. We also present our results on COCO 2017 Instance Segmentation and Object Detection Challenges.

\subsection{Implementation Details}
We re-implement Mask R-CNN and FPN based on Caffe \cite{jia2014caffe}. All pre-trained models we use in experiments are publicly available. We adopt image centric training \cite{girshick2015fast}. For each image, we sample $512$ region-of-interests (ROIs) with positive-to-negative ratio $1:3$. Weight decay is $0.0001$ and momentum is set to $0.9$. Other hyper-parameters slightly vary according to datasets and we detail them in respective experiments. Following Mask R-CNN, proposals are from an independently trained RPN \cite{lin2016feature,ren2015faster} for convenient ablation and fair comparison, \ie, the backbone is not shared with object detection/instance segmentation.

\begin{table*}[t]
	\centering \addtolength{\tabcolsep}{-3pt}
	\footnotesize
	\begin{tabular}{c|ccc|ccc|c}
		Method & AP$^{bb}$ & AP$_{50}^{bb}$ & AP$_{75}^{bb}$ & AP$_S^{bb}$ & AP$_M^{bb}$ & $AP_L^{bb}$ & Backbone \\
		\hline
		\hline
		Champion 2016 \cite{huang2017speed} & 41.6 & 62.3 & 45.6 & 24.0 & 43.9 & 55.2 & $2\times$ResNet-101 + $3\times$Inception-ResNet-v2\\
		RentinaNet \cite{Lin2017focal} & 39.1 & 59.1 & 42.3 & 21.8 & 42.7 & 50.2 & ResNet-101 \\
		Mask R-CNN \cite{He2017Mask}+FPN \cite{lin2016feature} & 38.2 & 60.3 & 41.7 & 20.1 & 41.1 & 50.2 & ResNet-101 \\
		Mask R-CNN \cite{He2017Mask}+FPN \cite{lin2016feature} & 39.8 & 62.3 & 43.4 & 22.1 & 43.2 & 51.2 & ResNeXt-101 \\
		\hline
		\hline
		PANet / PANet [ms-train] & 41.2 / 42.5 & 60.4 / 62.3 & 44.4 / 46.4 & 22.7 / 26.3 & 44.0 / 47.0 & 54.6 / 52.3 & ResNet-50 \\
		PANet / PANet [ms-train] & 45.0 / \bf 47.4 & 65.0 / \bf 67.2 & 48.6 / \bf 51.8 & 25.4 / \bf 30.1 & 48.6 / \bf 51.7 & 59.1 / \bf 60.0 & ResNeXt-101 \\
	\end{tabular}\vspace{0.1cm}
	\caption{{Comparison among PANet, winner of COCO 2016 object detection challenge, RentinaNet and Mask R-CNN on COCO {\it test-dev} subset in terms of box AP, where the latter three are baselines.}}\label{tab:maincoco_det}
	\vspace{-0.05in}
\end{table*}

\subsection{Experiments on COCO}
\paragraph{Dataset and Metrics}
COCO \cite{lin2014microsoft} dataset is among the most challenging ones for instance segmentation and object detection due to the data complexity. It consists of 115k images for training and 5k images for validation (new split of 2017). 20k images are used in {\it test-dev} and 20k images are used as {\it test-challenge}. Ground-truth labels of both {\it test-challenge} and {\it test-dev} are not publicly available. There are $80$ classes with pixel-wise instance mask annotation. We train our models on {\it train-2017} subset and report results on {\it val-2017} subset for ablation study. We also report results on {\it test-dev} for comparison. 

We follow the standard evaluation metrics, \ie, AP, AP$_{50}$, AP$_{75}$, AP$_S$, AP$_M$ and $AP_L$. The last three measure performance with respect to objects with different scales. Since our framework is general to both instance segmentation and object detection, we also train independent object detectors. We report mask AP, box ap AP$^{bb}$ of an independently trained object detector, and box ap AP$^{bbM}$ of the object detection branch trained in the multi-task fashion.

\vspace{-0.1in}
\paragraph{Hyper-parameters}
We take $16$ images in one image batch for training. The shorter and longer edges of the images are $800$ and $1000$, if not specially noted. For instance segmentation, we train our model with learning rate $0.02$ for 120k iterations and $0.002$ for another 40k iterations. For object detection, we train one object detector without the mask prediction branch. Object detector is trained for 60k iterations with learning rate $0.02$ and another 20k iterations with learning rate $0.002$. These parameters are adopted from Mask R-CNN and FPN without any fine-tuning.

\vspace{-0.1in}
\paragraph{Instance Segmentation Results}
We report performance of our PANet on {\it test-dev} for comparison, with and without multi-scale training. As shown in Table \ref{tab:maincoco}, our PANet with ResNet-50 trained on multi-scale images and tested on single-scale images already outperforms Mask R-CNN and Champion in 2016, where the latter used larger model ensembles and testing tricks \cite{he2016deep,li2017fully,dai2015instance,gidaris2015object,liu2015box,zagoruyko2016a}. Trained and tested with image scale $800$, which is same as that of Mask R-CNN, our method outperforms the single-model state-of-the-art Mask R-CNN with nearly $3$ points under the same initial models.

\vspace{-0.1in}
\paragraph{Object Detection Results}
Similar to the way adopted in Mask R-CNN, we also report bounding box results inferred from the box branch. Table \ref{tab:maincoco_det} shows that our method with ResNet-50, trained and tested on single-scale images, outperforms, by a large margin, all other single-model ones even using much larger ResNeXt-101 \cite{xie2016aggregated} as initial model. With multi-scale training and single-scale testing, our PANet with ResNet-50 outperforms Champion in 2016, which used larger model ensemble and testing tricks.

\begin{table*}[t]
\begin{center}
\addtolength{\tabcolsep}{-3.5pt}
{\footnotesize
\begin{tabular}{cccccccc|ccc|ccc}
MRB & RBL & MST & MBN & BPA & AFP & FF & HHD & AP/AP$^{bb}$/AP$^{bbM}$ & AP$_{50}$ & AP$_{75}$ & AP$_S$/AP$_S^{bb}$/AP$_S^{bbM}$ & AP$_M$/AP$_M^{bb}$/AP$_M^{bbM}$ & $AP_L$/$AP_L^{bb}$/$AP_L^{bbM}$  \\
\hline
\hline
\checkmark & - & - & - & - & - & - & - & 33.6 / 33.9 / ~~~-~~~  & 55.2 & 35.3 & ~~~-~~~ / 17.8 / ~~~-~~~ & ~~~-~~~ / 37.7 / ~~~-~~~ & ~~~-~~~ / 45.8 / ~~~-~~~ \\
 & \checkmark &  &  &  &  &  &  & 33.4 / 35.0 / 36.4 & 54.3 & 35.5 & 14.1 / 18.7 / 20.0 & 35.7 / 38.9 / 39.7 & 50.8 / 47.0 / 48.8 \\
 & \checkmark & \checkmark &  &  &  &  &  & 35.3 / 35.0 / 38.2 & 56.7 & 37.9 & 17.6 / 20.8 / 24.3 & 38.6 / 39.9 / 42.3 & 50.6 / 44.1 / 48.8  \\
  & \checkmark & \checkmark & \checkmark &  &  &  &  & 35.7 / 37.1 / 38.9 & 57.3 & 38.0 & 18.6 / 24.2 / 25.3 & 39.4 / 42.5 / 43.6 & 51.7 / 47.1 / 49.9 \\
  & \checkmark & \checkmark & \checkmark & \checkmark &  &  &  & 36.4 / 38.0 / 39.9 & 57.8 & 39.2 & 19.3 / 23.3 / 26.2 & 39.7 / 42.9 / 44.3 & 52.6 / 49.4 / 51.3 \\
  & \checkmark & \checkmark & \checkmark &  & \checkmark &  &  & 36.3 / 37.9 / 39.6 & 58.0 & 38.9 & 19.0 / 25.4 / 26.4 & 40.1 / 43.1 / 44.9 & 52.4 / 48.6 / 50.5 \\
  & \checkmark & \checkmark & \checkmark & \checkmark & \checkmark &  &  & 36.9 / 39.0 / 40.6 & 58.5 & 39.7 & 19.6 / 25.7 / \bf 27.0 & 40.7 / 44.2 / 45.7 & 53.2 / 49.5 / 52.1 \\
  & \checkmark & \checkmark & \checkmark & \checkmark & \checkmark & \checkmark &  & 37.6 / ~~~-~~~ / ~~~-~~~ & 59.1 & 40.6 & {\bf 20.3} / ~~~-~~~ / ~~~-~~~ & 41.3 / ~~~-~~~ / ~~~-~~~ & 53.8 / ~~~-~~~ / ~~~-~~~ \\
  & \checkmark & \checkmark & \checkmark & \checkmark & \checkmark & \checkmark & \checkmark & \bf 37.8 / 39.2 / 42.1 & \bf 59.4 & \bf 41.0 & 19.2 / \bf 25.8 / 27.0 & \bf 41.5 / 44.3 / 47.3 & \bf 54.3 / 50.6 / 54.1 \\
\hline
\hline
& &  & & & & & & \bf +4.4 / +4.2 / +5.7 & \bf +5.1 & \bf +5.5 & \bf +5.1 / +7.1 / +7.0 & \bf +5.8 / +5.4 / +7.6 & \bf +3.5 / +3.6 / +5.3
\end{tabular}\vspace{-0.1in}}
\end{center}
\caption{Performance in terms of mask AP, box ap AP$^{bb}$ of an independently trained object detector, and box ap AP$^{bbM}$ of the box branch trained with multi-task fashion on {\it val-2017}. Based on our re-implemented baseline (RBL), we gradually add multi-scale training (MST), multi-GPU synchronized batch normalization (MBN), bottom-up path augmentation (BPA), adaptive feature pooling (AFP), fully-connected fusion (FF) and heavier head (HHD) for ablation studies. MRB is short for Mask R-CNN result reported in the original paper. The last line shows total improvement compared with baseline RBL.}\label{tab:ablations}
\vspace{-0.1in}
\end{table*}

\vspace{-0.1in}
\paragraph{Component Ablation Studies}
First, we analyze importance of each proposed component. Besides {\it bottom-up path augmentation}, {\it adaptive feature pooling} and {\it fully-connected fusion}, we also analyze {\it multi-scale training}, {\it multi-GPU synchronized batch normalization} \cite{zhao2017pyramid,ioffe2015batch} and {\it heavier head}. For {\it multi-scale training}, we set longer edge to $1,400$ and the other to range from $400$ to $1,400$. We calculate mean and variance based on all samples in one batch across all GPUs, do not fix any parameters during training, and make all new layers followed by a batch normalization layer, when using {\it multi-GPU synchronized batch normalization}. The {\it heavier head} uses $4$ consecutive $3\times3$ convolutional layers shared by box classification and box regression, instead of two {\it fc} layers. It is similar to the head used in \cite{Lin2017focal} but the convolutional layers for box classification and box regression branches are not shared in their case.

Our ablation study from the baseline gradually to all components incorporated is conducted on {\it val-2017} subset and is shown in Table \ref{tab:ablations}. ResNet-50 \cite{he2016deep} is our initial model. We report performance in terms of mask AP, box ap AP$^{bb}$ of an independently trained object detector and box ap AP$^{bbM}$ of box branch trained in the multi-task fashion.

\vspace{0.05in}
\noindent {\bf 1) Re-implemented Baseline.~~} Our re-implemented Mask R-CNN performs comparable with the one described in original paper and our object detector performs better.

\vspace{0.05in}
\noindent {\bf 2) Multi-scale Training \& Multi-GPU Sync. BN.~~} These two techniques help the network to converge better and increase the generalization ability.

\vspace{0.05in}
\noindent {\bf 3) Bottom-up Path Augmentation.~~} With or without adaptive feature pooling, bottom-up path augmentation consistently improves mask AP and box ap AP$^{bb}$ by more than $0.6$ and $0.9$ respectively. The improvement on instances with large scale is most significant. This verifies usefulness of information from lower feature levels.

\vspace{0.05in}
\noindent {\bf 4) Adaptive Feature Pooling.~~} With or without bottom-up path augmentation, adaptive feature pooling consistently improves performance. The performance in all scales generally improves, in accordance with our observation that features in other layers are also useful in final prediction.

\vspace{0.05in}
\noindent {\bf 5) Fully-connected Fusion.~~} Fully-connected fusion aims at predicting masks with better quality. It yields $0.7$ improvement in terms of mask AP. It is general for instances at all scales.

\vspace{0.05in}
\noindent {\bf 6) Heavier Head.~~} Heavier head is quite effective for box ap AP$^{bbM}$ of bounding boxes trained with multi-task fashion. While for mask AP and independently trained object detector, the improvement is minor.

With all these components in PANet, improvement on mask AP is $4.4$ over baselines. Box ap AP$^{bb}$ of independently trained object detector increases $4.2$. They are significant. Small- and medium-size instances contribute most. Half improvement is from {\it multi-scale training} and {\it multi-GPU sync. BN}, which are effective strategies to help train better models. 


\vspace{-0.1in}
\paragraph{Ablation Studies on Adaptive Feature Pooling}\label{sec:afpabla}
We conduct ablation studies on adaptive feature pooling to find where to place the fusion operation and the most appropriate fusion operation. We place it either between ROIAlign and {\it fc1}, represented by ``fu.{\it fc1}{\it fc2}'' or between {\it fc1} and {\it fc2}, represented by ``{\it fc1}fu.{\it fc2}'' in Table \ref{tab:afpabla}. Similar settings are also applied to mask prediction branch. For feature fusing, max and sum operations are tested.

As shown in Table \ref{tab:afpabla}, adaptive feature pooling is not sensitive to the fusion operation. Allowing a parameter layer to adapt feature grids from different levels, however, is of great importance. We use max as fusion operation and use it behind the first parameter layer in our framework.

\begin{table}[t]
\centering \addtolength{\tabcolsep}{-1pt}
\footnotesize
\begin{tabular}{c|ccc|ccc}
Settings & AP & AP$_{50}$ & AP$_{75}$ & AP$^{bb}$ & AP$_{50}^{bb}$ & AP$_{75}^{bb}$  \\
\hline
\hline
baseline & 35.7 & 57.3 & 38 & 37.1 & 58.9 & 40.0\\
\hline
\hline
fu.{\it fc1}{\it fc2} & 35.7 & 57.2 & 38.2 & 37.3 & 59.1 & 40.1 \\
{\it fc1}fu.{\it fc2} & {\bf 36.3} & {\bf 58.0} & {\bf 38.9} & {\bf 37.9} & {\bf 60.0} & {\bf 40.7} \\
\hline
\hline
MAX & {\bf 36.3} & {\bf 58.0} & {\bf 38.9} & 37.9 & {\bf 60.0} & {\bf 40.7} \\
SUM & 36.2 & \bf 58.0 & 38.8 & {\bf 38.0} & 59.8 & \bf 40.7 \\
\end{tabular}\vspace{0.1cm}
\caption{Ablation study on adaptive feature pooling on {\it val-2017} in terms of mask AP and box ap AP$^{bb}$ of the independently trained object detector.}\label{tab:afpabla}
\vspace{-0.1in}
\end{table}

\vspace{-0.1in}
\paragraph{Ablation Studies on Fully-connected Fusion} 
We investigate performance with different ways to instantiate the augmented {\it fc} branch. We consider two aspects, \ie, the layer to start the new branch and the way to fuse predictions from the new branch and FCN. We experiment with creating new paths from $conv2$, $conv3$ and $conv4$, respectively. ``max", ``sum" and ``product" operations are used for fusion. We take our reimplemented Mask R-CNN with bottom-up path augmentation and adaptive feature pooling as the baseline. Corresponding results are shown in Table \ref{tab:ffabla}. They clearly show that staring from $conv3$ and taking sum for fusion produces the best results.

\begin{table}[t]
\centering \addtolength{\tabcolsep}{-1pt}
\footnotesize
\begin{tabular}{c|ccc|ccc}
Settings & AP & AP$_{50}$ & AP$_{75}$ & AP$_S$ & AP$_M$ & AP$_L$  \\
\hline
\hline
baseline & 36.9 & 58.5 & 39.7 & 19.6 & 40.7 & 53.2 \\
\hline
\hline
conv2 & 37.5 & {\bf 59.3} & 40.1 & {\bf 20.7} & 41.2 & {\bf 54.1} \\
conv3 & {\bf 37.6} & 59.1 & {\bf 40.6} & 20.3 & {\bf 41.3} & 53.8 \\
conv4 & 37.2 & 58.9 & 40.0 & 19.0 & 41.2 & 52.8 \\
\hline
\hline
PROD & 36.9 & 58.6 & 39.7 & 20.2 & 40.8 & 52.2 \\
SUM & {\bf 37.6} & {\bf 59.1} & {\bf 40.6} & {\bf 20.3} & {\bf 41.3} & {\bf 53.8} \\
MAX & 37.1 & 58.7 & 39.9 & 19.9 & 41.1 & 52.5 \\
\end{tabular}\vspace{0.1cm}
\caption{Ablation study on fully-connected fusion on {\it val-2017} in terms of mask AP.}\vspace{-0.1in}\label{tab:ffabla}
\end{table}


\vspace{-0.1in}
\paragraph{COCO 2017 Challenge}
With PANet, we participated in the COCO 2017 Instance Segmentation and Object Detection Challenges. Our framework reaches the $1^{st}$ place in Instance Segmentation task and the $2^{nd}$ place in Object Detection task without large-batch training. As shown in Tables \ref{tab:maskchallenge} and \ref{tab:boxchallenge}, compared with last year champions, we achieve $9.1\%$ absolute and $24\%$ relative improvement on instance segmentation. While for object detection, $9.4\%$ absolute and $23\%$ relative improvement is yielded.

\begin{table}[t]
\centering \addtolength{\tabcolsep}{-2pt}
\footnotesize
\begin{tabular}{c|ccc|ccc}
 & AP & AP$_{50}$ & AP$_{75}$ & AP$_S$ & AP$_M$ & AP$_L$  \\
\hline
\hline
Champion 2015 \cite{dai2015instance} & 28.4 & 51.6 & 28.1 & 9.4 & 30.6 & 45.6\\
Champion 2016 \cite{li2017fully} & 37.6 & 59.9 & 40.4 & 17.1 & 41.0 & 56.0 \\
Our Team 2017 & \bf 46.7 & \bf 69.5 & \bf 51.3 & \bf 26.0 & \bf 49.1 & \bf 64.0 \\
\hline
\hline
PANet baseline & 38.2 & 60.2 & 41.4 & 19.1 & 41.1 & 52.6 \\
+DCN \cite{Dai2017deformable}  & 39.5 & 62.0 & 42.8 & 19.8 & 42.2 & 54.7 \\
+testing tricks & 42.0 & 63.5 & 46.0 & 21.8 & 44.4 & 58.1  \\
+larger model & 44.4 & 67.0 & 48.5 & 23.6 & 46.5 & 62.2 \\
+ensemble & \bf 46.7 & \bf 69.5 & \bf 51.3 & \bf 26.0 & \bf 49.1 & \bf 64.0 \\
\end{tabular}\vspace{0.1cm}
\caption{Mask AP of COCO Instance Segmentation Challenge in different years on {\it test-dev}.}\label{tab:maskchallenge}
\vspace{-0.03in}
\end{table}

\begin{table}[t]
\centering \addtolength{\tabcolsep}{-2pt}
\footnotesize
\begin{tabular}{c|ccc|ccc}
 & AP$^{bb}$ & AP$_{50}^{bb}$ & AP$_{75}^{bb}$ & AP$_S^{bb}$ & AP$_M^{bb}$ & AP$_L^{bb}$  \\
\hline
\hline
Champion 2015 \cite{he2016deep} & 37.4 & 59.0 & 40.2 & 18.3 & 41.7 & 52.9 \\
Champion 2016 \cite{huang2017speed} & 41.6 & 62.3 & 45.6 & 24.0 & 43.9 & 55.2 \\
Our Team 2017 & \bf 51.0 & \bf 70.5 & \bf 55.8 & \bf 32.6 & \bf 53.9 & \bf 64.8 \\
\end{tabular}\vspace{0.1cm}
\caption{Box AP of COCO Object Detection Challenge in different years on {\it test-dev}.}\vspace{-0.1in}\label{tab:boxchallenge}
\vspace{-0.02in}
\end{table}

\begin{figure*}[bpt]
\centering
\begin{tabular}{@{\hspace{0mm}}c}
\includegraphics[width=0.85\linewidth]{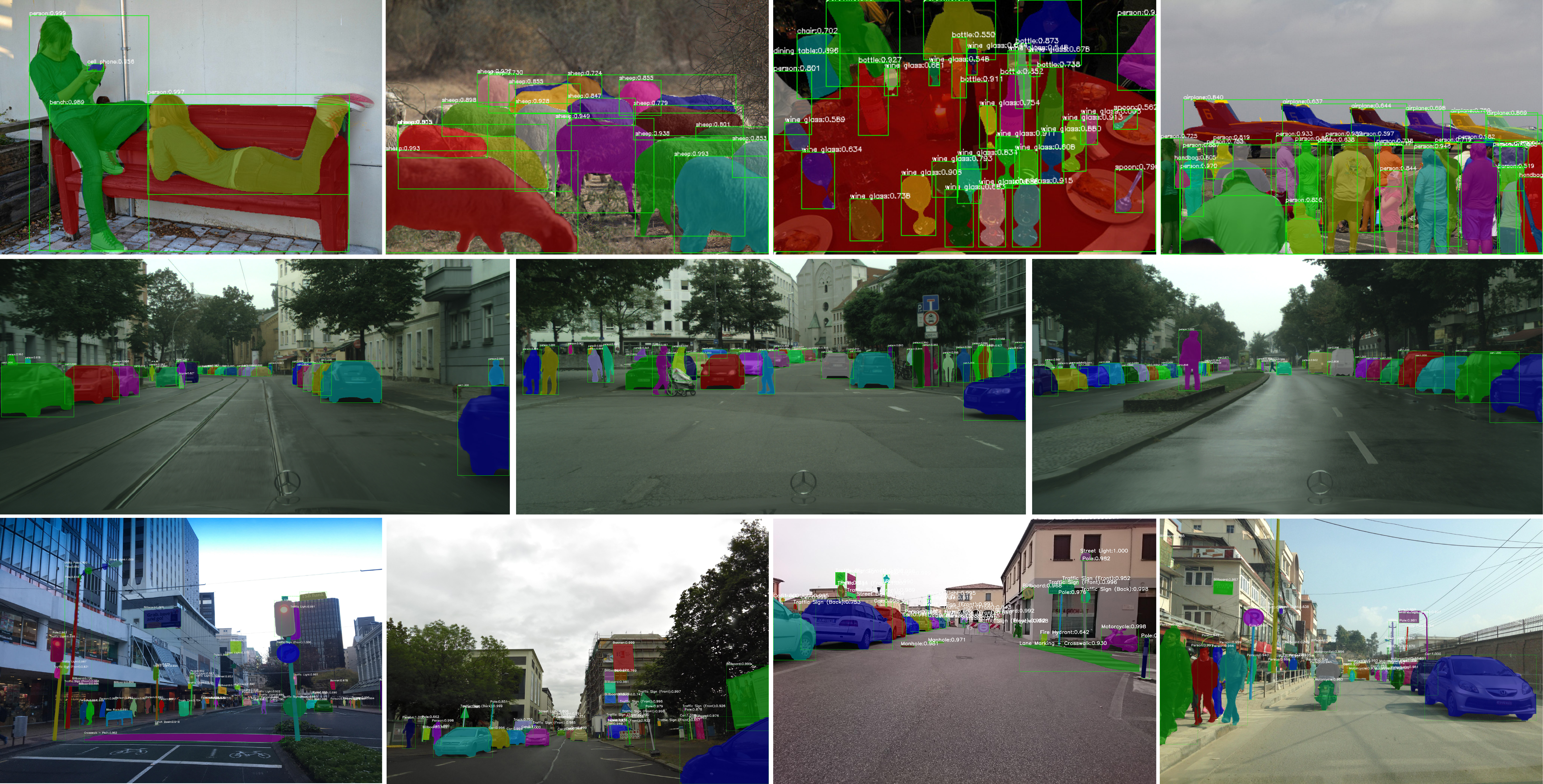}
\\
\end{tabular}
\caption{Images in each row are visual results of our model on COCO {\it test-dev}, Cityscapes {\it test} and MVD {\it test}, respectively.}
\label{fig:visual_coco}
\vspace{-0.05in}
\end{figure*}

The top performance comes with a few more details in PANet. First, we use deformable convolutions where DCN \cite{Dai2017deformable} is adopted. The common testing tricks \cite{he2016deep,li2017fully,dai2015instance,gidaris2015object,liu2015box,zagoruyko2016a}, such as multi-scale testing, horizontal flip testing, mask voting and box voting, are adopted. For multi-scale testing, we set the longer edge to $1,400$ and the other ranges from $600$ to $1,200$ with step $200$. Only $4$ scales are used. Second, we use larger initial models from publicly available ones. We use 3 ResNeXt-101 ($64\times4d$) \cite{xie2016aggregated}, 2 SE-ResNeXt-101 ($32\times4d$) \cite{hu2017senet}, 1 ResNet-269 \cite{zeng2016crafting} and 1 SENet \cite{hu2017senet} as ensemble for bounding box and mask generation. Performance with different larger initial models are similar. One ResNeXt-101 ($64\times4d$) is used as the base model to generate proposals. We train these models with different random seeds, with and without balanced sampling \cite{shen2016relay} to enhance diversity between models. Detection results we submitted are acquired by tightening instance masks. We show a few visual results in Figure \ref{fig:visual_coco} -- most of our predictions are with high quality.

\begin{table*}[t]
\centering \addtolength{\tabcolsep}{-1pt}
\footnotesize
\begin{tabular}{c|c|cc|cccccccc}
Methods & AP [val] & AP & AP$_{50}$ & person & rider & car & truck & bus & train & motorcycle & bicycle  \\
\hline
\hline
SGN \cite{liu2017sgn}& 29.2 & 25.0 & 44.9 & 21.8 & 20.1 & 39.4 & 24.8 & 33.2 & 30.8 & 17.7 & 12.4\\
Mask R-CNN [fine-only] \cite{He2017Mask} & 31.5 & 26.2 & 49.9 & 30.5 & 23.7 & 46.9 & 22.8 & 32.2 & 18.6 & 19.1 & 16.0 \\
SegNet & - & 29.5 & 55.6 & 29.9 & 23.4 & 43.4 & 29.8 & 41.0 & \bf 33.3 & 18.7 & 16.7 \\
Mask R-CNN [COCO] \cite{He2017Mask} & 36.4 & 32.0 & 58.1 & 34.8 & 27.0 & 49.1 & 30.1 & 40.9 & 30.9 & 24.1 & 18.7 \\
\hline
\hline
PANet [fine-only] & 36.5 & 31.8 & 57.1 & 36.8 & 30.4 & 54.8 & 27.0 & 36.3 & 25.5 & 22.6 & 20.8 \\
PANet [COCO] & \bf 41.4 & \bf 36.4 & \bf 63.1 & \bf 41.5 & \bf 33.6 & \bf 58.2 & \bf 31.8 & \bf 45.3 & 28.7 & \bf 28.2 & \bf 24.1 \\
\end{tabular}\vspace{0.1cm}
\caption{Results on Cityscapes {\it val} subset, denoted as AP [val], and on Cityscapes {\it test} subset, denoted as AP.}\label{tab:cityscapes}
\vspace{-0.1in}
\end{table*}

\begin{table}[t]
\centering \addtolength{\tabcolsep}{-1pt}
\footnotesize
\begin{tabular}{c|cc}
Methods & AP & AP$_{50}$ \\
\hline
\hline
our re-implement  & 33.1 & 59.1 \\
our re-implement + MBN  & 34.6 & 62.4 \\
PANet & \bf 36.5 & \bf 62.9 \\
\end{tabular}\vspace{0.1cm}
\caption{Ablation study results on Cityscapes {\it val} subset. Only fine annotations are used for training. MBN is short for multi-GPU synchronized batch normalization.}\label{tab:cityscapes_abl}
\vspace{-0.1in}
\end{table}

\subsection{Experiments on Cityscapes}
\label{sec:cityscapes}
\paragraph{Dataset and Metrics}
Cityscapes \cite{cordts2016the} contains street scenes captured by car-mounted cameras. There are $2,975$ training images, $500$ validation images and $1,525$ testing images with fine annotations. Another $20$k images are with coarse annotations, excluded for training. We report our results on {\it val} and secret {\it test} subsets. $8$ semantic classes are annotated with instance masks. Each image is with size $1024\times2048$. We evaluate results based on AP and AP$_{50}$.

\vspace{-0.1in}
\paragraph{Hyper-parameters}
We use the same set of hyper-parameters as in Mask R-CNN \cite{He2017Mask} for fair comparison. Specifically, we use images with shorter edge randomly sampled from $\{800, 1024\}$ for training and use images with shorter edge $1024$ for inference. No testing tricks or DCN is used. We train our model for $18$k iterations with learning rate $0.01$ and another $6$k iterations with learning rate $0.001$. $8$ images ($1$ image per GPU) are in one image batch. ResNet-50 is taken as the initial model on this dataset.

\vspace{-0.1in}
\paragraph{Results and Ablation Study}
We compare with state-of-the-arts on {\it test} subset in Table \ref{tab:cityscapes}. Trained on ``fine-only" data, our method outperforms Mask R-CNN with ``fine-only" data by $5.6$ points. It is even comparable with Mask R-CNN pre-trained on COCO. By pre-training on COCO, we outperform Mask R-CNN with the same setting by $4.4$ points. Visual results are shown in Figure \ref{fig:visual_coco}.

Our ablation study to analyze the improvement on {\it val} subset is given in Table \ref{tab:cityscapes_abl}. Based on our re-implemented baseline, we add multi-GPU synchronized batch normalization to help network converge better. It improves the accuracy by $1.5$ points. With our full PANet, the performance is further boosted by $1.9$ points. 

\begin{table}[t]
\centering \addtolength{\tabcolsep}{-4pt}
\footnotesize
\begin{tabular}{c|cc|cc}
Methods & AP [test] & AP$_{50}$ [test] & AP [val] & AP$_{50}$ [val] \\
\hline
\hline
UCenter-Single \cite{liu2017lsun2017} & - & - & 22.8 & 42.5 \\
UCenter-Ensemble \cite{liu2017lsun2017} & 24.8 & 44.2 & 23.7 & 43.5 \\
\hline
PANet  & - & - & 23.6 & 43.3 \\
PANet [test tricks] & \bf 26.3 & \bf 45.8 & \bf 24.9 & \bf 44.7 \\
\end{tabular}\vspace{0.1cm}
\caption{Results on MVD {\it val} subset and {\it test} subset.}\label{tab:mvd}
\vspace{-0.1in}
\end{table}

\subsection{Experiments on MVD}
MVD \cite{Neuhold2017the} is a relatively new and large-scale dataset for instance segmentation. It provides $25,000$ images on street scenes with fine instance-level annotations for $37$ semantic classes. They are captured from several countries using different devices. The content and resolution vary greatly. We train our model on {\it train} subset with ResNet-50 as initial model and report performance on {\it val} and secret {\it test} subsets in terms of AP and AP$_{50}$.

We present our results in Table \ref{tab:mvd}. Compared with UCenter \cite{liu2017lsun2017} -- winner on this dataset in LSUN 2017 instance segmentation challenge, our PANet with one ResNet-50 tested on single-scale images already performs comparably with the ensemble result with pre-training on COCO. With multi-scale and horizontal flip testing, which are also adopted by UCenter, our method performs even better. Qualitative results are illustrated in Figure \ref{fig:visual_coco}.

\vspace{-0.05in}
\section{Conclusion}
We have presented our PANet for instance segmentation. We designed several simple and yet effective components to enhance information propagation in representative pipelines. We pool features from all feature levels and shorten the distance among lower and topmost feature levels for reliable information passing. Complementary path is augmented to enrich feature for each proposal. Impressive results are produced. Our future work will be to extend our method to video and RGBD data.

\section*{Acknowledgements}
We would like to thank Yuanhao Zhu, Congliang Xu and Qingping Fu in SenseTime for technical support.

\section*{Appendix}
\begin{appendices}  

\section{Training Details and Strategy of Generating Anchors on Cityscapes and MVD.}

On Cityscapes \cite{cordts2016the}, training hyper-parameters are adopted from Mask R-CNN \cite{He2017Mask} and described in Section \ref{sec:cityscapes}. RPN anchors span $5$ scales and $3$ aspect ratios following \cite{He2017Mask,lin2016feature}. While on MVD \cite{Neuhold2017the}, we adopt training hyper-parameters from the winning entry \cite{liu2017lsun2017}. We train our model with learning rate $0.02$ for 60k iterations and $0.002$ for another 20k iterations. We take $16$ images in one image batch for training. We set the longer edge of the input image to $2400$ pixels and the other ranges from $600$ to $2000$ pixels for multi-scale training. Scales $\{1600, 1800, 2000\}$ are adopted for multi-scale testing. The RPN anchors span $7$ scales, \ie, $\{8^2, 16^2, 32^2, 64^2, 128^2, 256^2, 512^2\}$, and $5$ aspect ratios, \ie, $\{0.2, 0.5, 1, 2, 5\}$. RPN is trained with the same scales as those of object detection/instance segmentation network training.

\section{Details on Implementing Multi-GPU Synchronized Batch Normalization.}

We implement multi-GPU batch normalization on Caffe \cite{jia2014caffe} and OpenMPI. Given $n$ GPUs and training samples in batch $B$, we first split training samples evenly into $n$ sub-batches, each is denoted as $b_i$, assigned to one GPU. On each GPU, we calculate means $\mu_i$ based on samples in $b_i$. AllReduce operation is then applied to gather all $\mu_i$ across all GPUs to get the mean $\mu_B$ of entire batch $B$. $\mu_B$ is broadcast to all GPUs. We then calculate temporary statistics on each GPU independently and apply the AllReduce operation to produce the variance $\sigma^2_B$ of entire batch $B$. $\sigma^2_B$ is also broadcast to all GPUs. As a result, each GPU has the statistics calculated on all training samples in $B$. We then perform  normalization $y_m=\gamma\frac{x_m - \mu_B}{\sqrt{\sigma^2_B + \epsilon}}+\beta$ as in \cite{ioffe2015batch} for each training sample. In backward operations, AllReduce operation is similarly applied to gather information from all GPUs for gradient calculation.

\begin{figure}[bpt]
\centering
\begin{tabular}{@{\hspace{0mm}}c}
\includegraphics[width=0.95\linewidth]{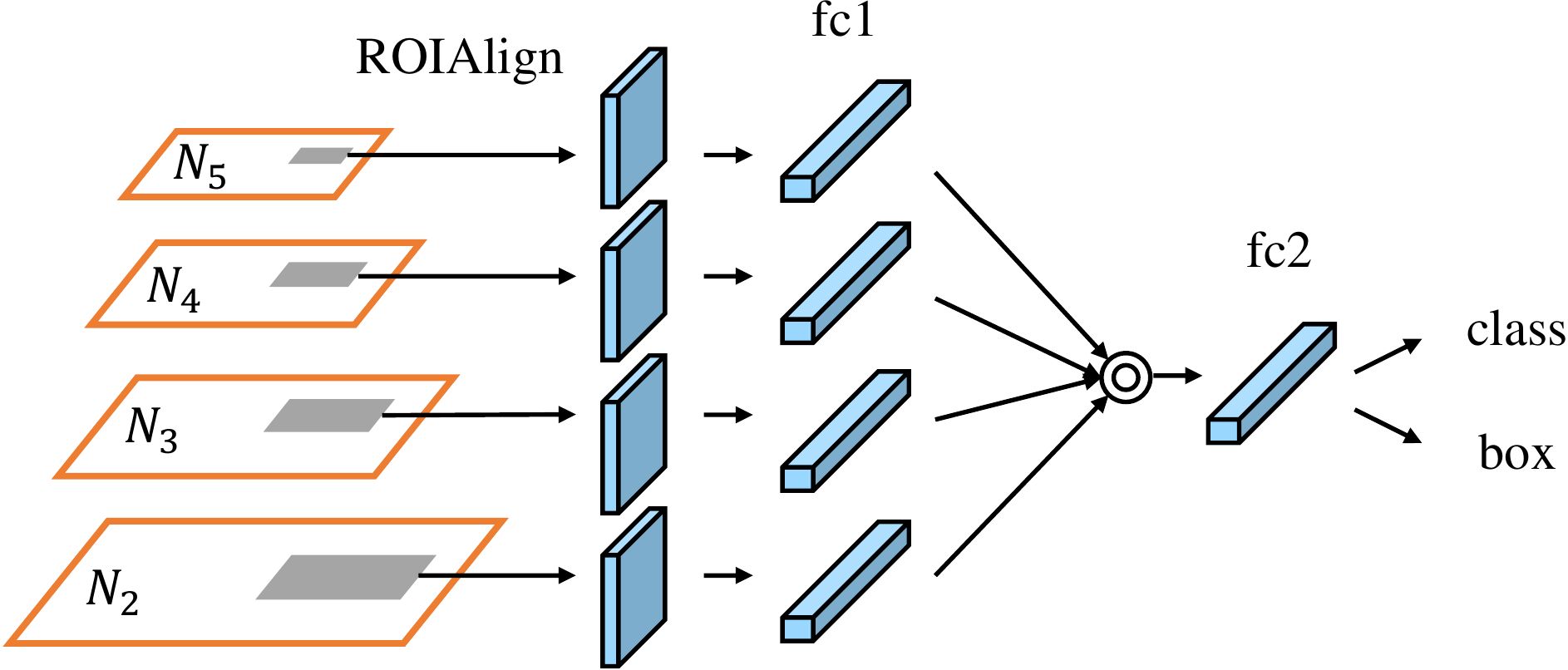}
\\
\end{tabular}
\caption{Illustration of adaptive feature pooling on box branch.}
\label{fig:ada}
\end{figure}

\end{appendices}

{\small
\bibliographystyle{ieee}
\bibliography{egbib}
}

\end{document}